\documentclass{article}

\usepackage{microtype}
\usepackage{graphicx}
\usepackage{graphicx}
\usepackage{subfig}
\usepackage{booktabs} % for professional tables
\usepackage{microtype}
\usepackage{hyperref}

\usepackage[accepted]{icml2025}

\usepackage{amsmath}
\usepackage{amssymb}
\usepackage{mathtools}
\usepackage{amsthm}

\usepackage[capitalize,noabbrev]{cleveref}

\theoremstyle{plain}

\theoremstyle{definition}

\theoremstyle{remark}

\usepackage[textsize=tiny]{todonotes}

\icmltitlerunning{SayAnything: Audio-Driven Lip Synchronization with Conditional Video Diffusion}

\begin{document}
\begin{sloppypar}
\twocolumn[

\icmltitle{SayAnything: Audio-Driven Lip Synchronization with Conditional Video Diffusion}




\begin{icmlauthorlist}
\icmlauthor{Junxian Ma}{ucas,rb,ntu}
\icmlauthor{Shiwen Wang}{ucas,rb}
\icmlauthor{Jian Yang}{rb}
\icmlauthor{Junyi Hu}{pku,rb}
\icmlauthor{Jian Liang}{rb}\\
\icmlauthor{Guosheng Lin}{ntu}
\icmlauthor{Jingbo Chen}{ucas}
\icmlauthor{Kai Li}{cityu,ucas}
\icmlauthor{Yu Meng}{ucas}
\end{icmlauthorlist}

\icmlaffiliation{ucas}{University of Chinese Academy of Sciences}
\icmlaffiliation{rb}{RightBrain.AI}
\icmlaffiliation{pku}{Peking University}
\icmlaffiliation{ntu}{Nanyang Technological University}
\icmlaffiliation{cityu}{City University of Hong Kong}

\icmlcorrespondingauthor{Junxian Ma}{majunxian20@mails.ucas.ac.cn}

\icmlkeywords{Audio-Driven Generation, Video Diffusion Models, Lip Synchronization, Video Editing}

\vskip 0.3in

% \twocolumn[{%
{
\renewcommand\twocolumn[1][]{#1}%
% \maketitle
\vspace{-8mm}
\captionsetup{type=figure}
\begin{center}
    \includegraphics[width=\linewidth]{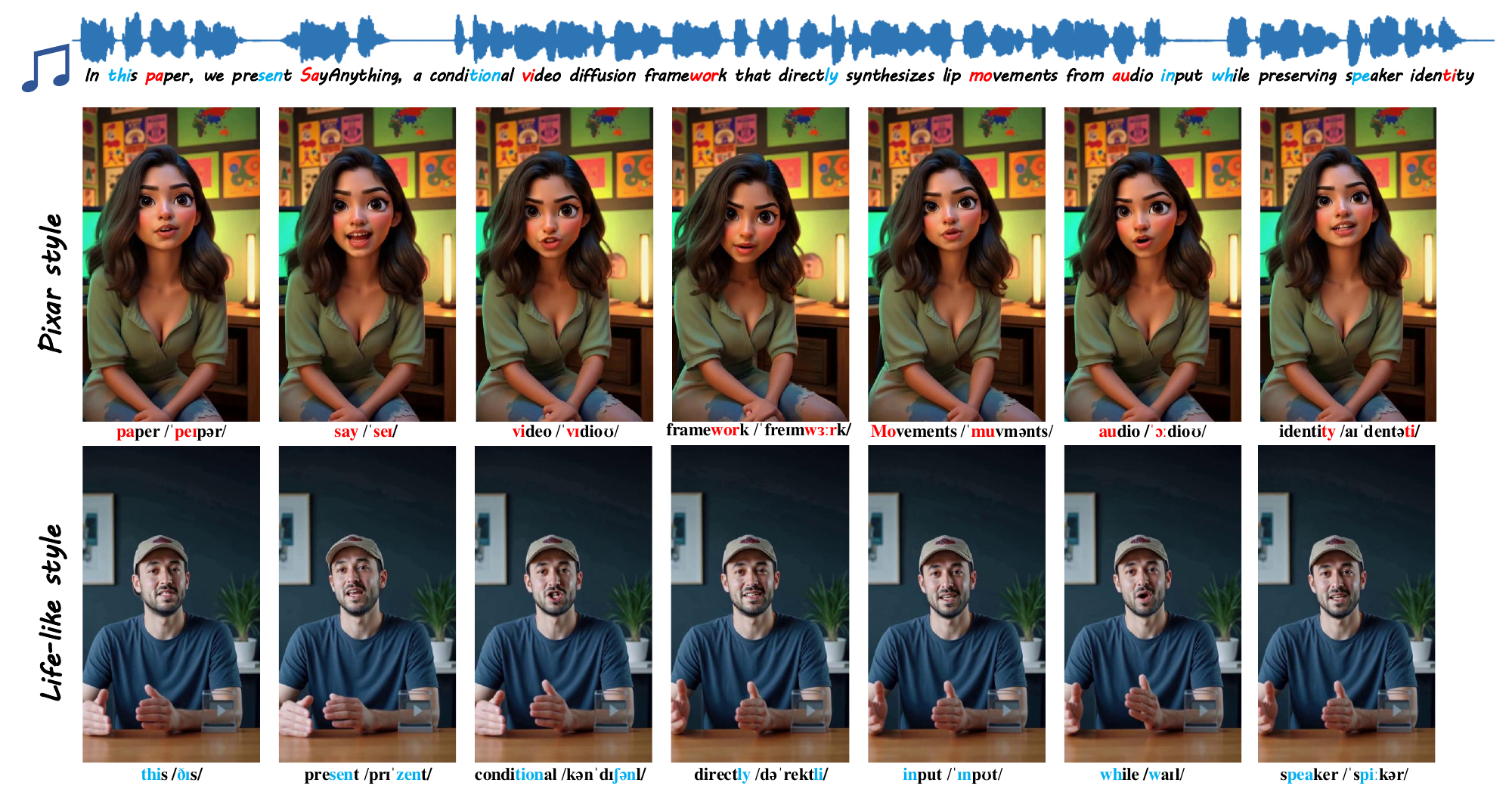}
    \caption{SayAnything performs audio-driven lip synchronization through video editing, demonstrating zero-shot generalization to in-the-wild and various style domains without fine-tuning. Our fusion scheme eliminates the dependency on additional supervision signals like SyncNet for lip synchronization. 
    % More video results are available in the supplementary materials.
    } %% wsw标题写附录合适吗？
    \label{fig:teaser}
    % \includegraphics[width=\linewidth]{figures/tessar}
    %     \caption{Automatic artistic typography generation results incorporating design elements. The first and third rows display the design elements, while the second and fourth rows show the corresponding typography outputs. The results demonstrate the integration of semantic elements and background color splatter.}
    % \label{fig:teaser}
    \end{center}%
}
]



\printAffiliationsAndNotice{Work done while the first author was an intern at RightBrain.AI, Beijing} % leave blank if no need to mention equal contribution
% \begin{abstract}
% This document provides a basic paper template and submission guidelines.
% Abstracts must be a single paragraph, ideally between 4--6 sentences long.
% Gross violations will trigger corrections at the camera-ready phase.
% \end{abstract}
% \input{icml2024/sec/tessar}
% \begin{figure*}[!htp]
%     \centering
%     \includegraphics[width=\textwidth]{icml2024/figures/intro.pdf}
%     \caption{SayAnything performs audio-driven lip synchronization through video editing, demonstrating zero-shot generalization to in-the-wild and various style domains without fine-tuning. Our fusion scheme eliminates the dependency on additional supervision signals like SyncNet for lip synchronization. More video results are available in the supplementary materials.}
%     \label{fig:teaser}
% \end{figure*}

\begin{abstract}
Recent advances in diffusion models have led to significant progress in audio-driven lip synchronization. However, existing methods typically rely on constrained audio-visual alignment priors or multi-stage learning of intermediate representations to force lip motion synthesis. This leads to complex training pipelines and limited motion naturalness. In this paper, we present SayAnything, a conditional video diffusion framework that directly synthesizes lip movements from audio input while preserving speaker identity. Specifically, we propose three specialized modules, including an identity preservation module, an audio guidance module, and an editing control module. Our novel design effectively balances different condition signals in the latent space, enabling precise control over appearance, motion, and region-specific generation without requiring additional supervision signals or intermediate representations. Extensive experiments demonstrate that SayAnything generates highly realistic videos with improved lip-teeth coherence, enabling unseen characters to \textbf{say anything} while effectively generalizing to animated characters.
\end{abstract}
\vspace{-1cm}

\section{Introduction}

Audio-driven lip synchronization aims to generate synchronized lip movements in videos based on input audio while preserving the speaker's identity and appearance. This task has significant applications in video dubbing, virtual avatars, and live-streaming platforms.

In the field of lip synchronization, GAN-based approaches~\citep{guan2023stylesync,su2024audio} remain the dominant paradigm. However, these methods face two major limitations. First, they lack stability and diversity in motion generation, resulting in unnatural visual effects. Second, the inherent training instability and mode collapse issues make them difficult to scale to diverse datasets~\citep{heusel2017gans}, limiting their practical applications.

With the success of diffusion models~\citep{ho2020denoising,song2020score} in image and video generation~\citep{blattmann2023stable,rombach2022high}, audio-driven facial animation has emerged as a promising research direction. Recent work leveraging diffusion models for lip synchronization encounters a fundamental challenge where models favour visual conditions over audio signals during generation. Existing approaches~\citep{mukhopadhyay2024diff2lip,li2024latentsync,shen2023difftalk,zhong2024high} force models to learn audio-visual correspondence through additional supervision from Syncnet~\citep{chung2017out} as lip experts~\citep{mukhopadhyay2024diff2lip,li2024latentsync} or intermediate representations like landmarks~\citep{shen2023difftalk,zhong2024high}. While this mitigates the challenge, the strong dependency on visual priors from lip experts and intermediate representations significantly constrains model generalizability and motion dynamics.

To address these limitations, we present SayAnything, an end-to-end framework that directly establishes audio-visual correlations in the latent space without requiring additional supervision signals or intermediate representations. Specifically, we design a multi-modal condition fusion scheme, incorporating three specialized modules: 1) For reference images, we design an identity preservation module with an efficient encoder ID-Guider that extracts identity features while controlling their influence on lip motion styles, 2) For videos, we propose an editing control module with an adaptive masking strategy that regulates the lip editing region and eliminates the dependency of lip movements on surrounding pixel motion patterns, and 3) For audio inputs, we develop an audio guidance module that enhances the control of weakly-correlated audio features over motion generation. As shown in \cref{fig:teaser}, our method demonstrates effective zero-shot generalization to both real-world and animated characters. To the best of our knowledge, this is the first work to leverage Stable Video Diffusion (SVD)~\citep{blattmann2023stable} for lip synchronization.

Our main contributions can be summarized as follows:
\begin{itemize}
   \item We propose an end-to-end video diffusion framework for audio-driven lip synchronization that eliminates the need for additional supervision signals or intermediate representations.
   
   \item We design a novel multi-modal condition fusion scheme that effectively balances audio-driven control and portrait preservation, incorporating identity preservation, audio guidance, and editing control modules.
   
   \item Both extensive experiments and user studies demonstrate that our approach achieves superior performance in terms of generation quality, temporal consistency, and lip motion flexibility compared to existing methods. Furthermore, our framework shows strong generalization capability to in-the-wild videos and animated characters.
\end{itemize}

\begin{figure*}[!ht]
    \centering
    \vspace{5mm}
    \includegraphics[width=0.9\textwidth]{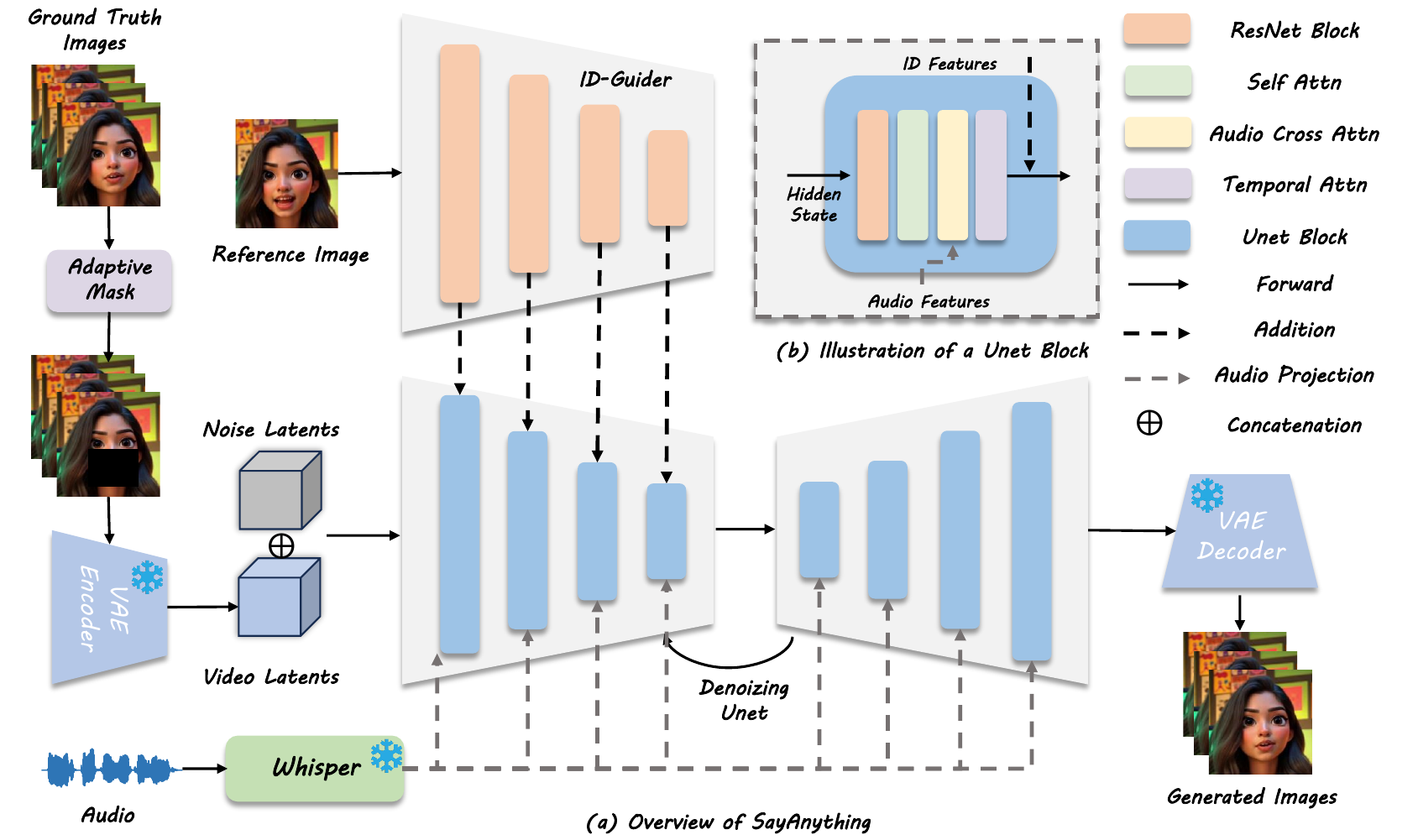}
    \caption{(a) Overview of SayAnything architecture for lip synchronization. The denoising UNet takes noisy latents as input, concatenated with video latents obtained from masked video through VAE encoding. The reference image is processed by ID-Guider to produce multi-scale ID features, which are injected as residual signals into the denoising UNet. Audio features from Whisper are fused through cross-attention layers in the denoising process. (b) A typical UNet block, consisting of ResNet block, Self Attention, Audio Cross Attention, and Temporal Attention.}
    \label{fig:overview}
\end{figure*}

\section{Related Work}
\subsection{Video Diffusion Models}

Diffusion generative models have recently gained prominence as a leading approach for vision generation tasks, with notable contributions from several studies~\citep{sohl2015deep, song2019generative, ho2020denoising, song2021scorebased, ruiz2024lane}. In the realm of video generation, diffusion models have also seen significant advancements, with various methods being proposed~\cite{videoworldsimulators2024, polyak2024movie, opensora2024, chen2024videocrafter2, blattmann2023stable, jta+22, jwc+22, arh+23, yca+23, xyg+24, rma+23, zhou2023magicvideoefficientvideogeneration}. For instance, the early diffusion-based video generation model VDM was introduced by~\cite{jta+22}. ImagenVideo~\cite{jwc+22} employs a cascaded diffusion process~\cite{ho2022cascaded} to generate high-resolution videos. MagicVideo~\cite{zhou2023magicvideoefficientvideogeneration} leverages the architecture of Latent Diffusion Models~\cite{rombach2022high} for video generation, utilizing a low-dimensional latent embedding space defined by a pre-trained variational auto-encoder (VAE). The text-to-video model SORA~\cite{videoworldsimulators2024} is capable of generating high-quality videos up to a minute long based on user prompts. Movie Gen~\cite{polyak2024movie}, developed by Meta, offers a suite of foundation models that can produce high-quality, high-resolution videos with synchronized audio in various aspect ratios. In the open-source domain, recent foundational video generation models include OpenSora~\cite{opensora2024}, VideoCrafter-2~\cite{chen2024videocrafter2}, and Stable Video Diffusion~\cite{blattmann2023stable}.

\subsection{Video Editing}

As diffusion-based video generation continues to advance, the ability to edit videos using text inputs has become a focal point of research. One notable contribution is Tune-A-Video~\cite{wgw+23}, which introduces the idea of one-shot video editing tuning. This method converts the spatial self-attention layers of a text-to-image (T2I) StableDiffusion model into sparse-casual attention layers. However, approaches like Tune-A-Video come with high fine-tuning costs.
To address this, SimDA~\cite{zqh+23} proposes a more parameter-efficient fine-tuning method to extend T2I StableDiffusion to text-to-video (T2V) generation. It achieves this by optimizing an adapter composed of two learnable fully connected (FC) layers. Another approach, Fairy~\cite{bcx+23}, focuses on parallelized frame sampling and keyframe editing using T2I diffusion models~\cite{bhe23} in a single forward pass.
Other works explore different conditions for video editing. For example, Gen-1~\cite{pjp+23} incorporates depth maps~\cite{RanftlLHSK22} to provide sequential depth guidance, although this method is computationally expensive due to the need to fine-tune all parameters. MagicEdit~\cite{jhj+23} freezes the pre-trained parameters of a T2I model and fine-tunes new temporal layers on video data, similar to AnimateDiff~\cite{yca+23}. It then integrates a ControlNet~\cite{zra23} trained for conditional image generation.
Ground-A-Video~\cite{hj23} leverages both depth maps and bounding boxes to provide spatial control over target regions using an inflated ControlNet~\cite{zra23}. Finally, VideoComposer~\cite{WangYZCWZSZZ23} expands the range of controllable conditions and allows simultaneous control with multiple inputs, including sketch images~\cite{0002LYH00P021}, depth maps~\cite{RanftlLHSK22}, and motion vectors~\cite{VadakitalDLTLR22}.
\vspace{-0.2cm}

\subsection{Audio-Driven Talking Head Generation}
Recently, GAN-based methods~\cite{prajwal2020lip, zhang2023dinet, zhong2023identity, wang2023lipformer, tan2024style2talker, tan2025edtalk, yang2024ladtalk, zhang2024musetalk, cheng2022videoretalking} have emerged as an essential research area in talking head generation, but they struggle to represent detailed texture making unremarkable video performance. On the other hand, with the recent rise of diffusion model in the AIGC community, numerous works~\cite{xu2024hallo, xu2024vasa, tian2025emo, chen2024echomimic, zhu2024infp} have been proposed to synthesize talking head video via sound and single portrait conditions and achieved remarkable performance in lip synchronization and visual effect. However, single portrait animation has fewer application scenarios than video dubbing. To this end, Diff2lip~\cite{mukhopadhyay2024diff2lip}  and LatentSync~\cite{li2024latentsync} are proposed, which both introduce a lip expert~\cite{prajwal2020lip} to offer strong audio-visual alignment in pixel space. While this expert benefits from the improvement of lip synchronization, the life-like prior that it contains hinders the variety of input video types. To this end, we propose SayAnything, which directly drives lip synchronization by leveraging the general visual priors from stable video diffusion to enable talking head generation across various styles.

\section{Method}
In this section, we will introduce our method, SayAnything. First, we will provide an overview of the framework and the preliminaries of SVD~\citep{blattmann2023stable}. Then, we will detail our multi-modal condition fusion scheme.

\subsection{Overview}

\paragraph{Stable Video Diffusion.} SVD is a high-quality and commonly used image-to-video generation model. Given a reference image $I$, SVD will generate a video frame sequence $x = \{x_0, x_1, ..., x_{N-1}\}$ of length N, starting with $I$. The sampling of SVD is conducted using a latent denoising diffusion process. At each denoising step, a conditional 3D UNet $\Phi_\theta$ is used to iteratively denoise this sequence:
\begin{equation}
\setlength{\abovedisplayskip}{3pt}
z^0 = \Phi_\theta(z_t, t, I)
\setlength{\belowdisplayskip}{3pt}
\end{equation}
where $z_t$ is the latent representation of $x_t$ and $z^0$ is the prediction of $z_0$. There are two conditional injection paths for the reference image $I$: (1) It is embedded into tokens by the CLIP~\citep{radford2021learning} image encoder and injected into the diffusion model through a cross-attention mechanism; (2) It is encoded into a latent representation by the VAE~\citep{kingma2013auto} encoder of the latent diffusion model~\citep{rombach2022high}, and concatenated with the latent of each frame in channel dimension.
SVD follows the EDM-preconditioning~\citep{karras2022elucidating} framework, which parameterizes the learnable denoiser 
\(\Phi_\theta\) as
\begin{small}
\begin{align}
\setlength{\abovedisplayskip}{0pt}
& \Phi_\theta\bigl(z_t,\,t,\,I;\sigma\bigr) 
= c_{skip}(\sigma)\,z_t \\
\nonumber
&\quad +\, c_{out}(\sigma)\,
F_\theta\Bigl(
  c_{in}(\sigma)\,z_t,\,
  t,\,
  I;\,
  c_{noise}(\sigma)
\Bigr).
\label{eq:edm_param}
\setlength{\belowdisplayskip}{0pt}
\end{align}
\end{small}
where \(\sigma\) is the noise level, and \(F_\theta\) is the network to be trained. 
\(c_{skip},\, c_{out},\, c_{in},\, \text{and}\; c_{noise}\) are preconditioning hyper-parameters.
\noindent
\(\Phi_\theta\) is trained via a denoising score matching (DSM)~\citep{song2019generative} objective:
\begin{equation}
\setlength{\abovedisplayskip}{3pt}
\mathbb{E}_{\substack{z_0,\,t\\n\,\sim\,\mathcal{N}(0,\sigma^2)}}
\Bigl[
  \lambda_\sigma
  \,\bigl\|
    \Phi_\theta\bigl(z_0 + n,\, t,\, I\bigr)
    \;-\;
    z_0
  \bigr\|_2^2
\Bigr].
\label{eq:edm_dsm}
\setlength{\belowdisplayskip}{3pt}
\end{equation}
\paragraph{SayAnything.}
Given a reference video $V_m$ and an audio $a$, our objective is to edit the lip motion of the person in $V_m$ so it aligns with $a$. 
To achieve this, three key elements are involved:
(1)~identity preservation via a reference image $\mathcal{I}_r$,
(2)~driving the spatial configuration of the lips based on the audio $a$,
and 
(3)~maintaining overall visual coherence by specifying an editable region in the video through masking.

To ensure broad generalization, our method inherits as many structures and parameters as possible from a pretrained stable video diffusion model, fully exploiting its robust visual priors. We design a unified, efficient multi-modal fusion scheme to balance these three conditioning signals—comprising an efficient identity preservation module, an 
editing control module for region edits, and an attention-based audio guidance module.

Specifically, we adapt SVD's denoiser $\Phi_{\theta}$ to fuse $\mathcal{I}_r$, $a$, and $V_m$ in a unified manner.
Below is our conditional denoiser, which replaces the original single-condition path:
\begin{small}
\begin{align}
\setlength{\abovedisplayskip}{0pt}
&\Phi_\theta\bigl(z_t,\,t,\,\mathcal{I}_r,\,a,\,V_m;\;\sigma\bigr)
~=~
c_{\mathrm{skip}}(\sigma)\,z_t 
\nonumber\\[-3pt]
&\quad~+\;
c_{\mathrm{out}}(\sigma)\;
F_\theta\Bigl(
    c_{\mathrm{in}}(\sigma)\,z_t,\;
    t,\;
    \mathcal{I}_r,\;
    a,\;
    V_m;\;
    c_{\mathrm{noise}}(\sigma)
\Bigr).
\setlength{\belowdisplayskip}{0pt}
\label{eq:ours_denoiser}
\end{align}
\end{small}
By jointly injecting the identity signal $\mathcal{I}_r$, the audio $a$, and the masked-video prior $V_m$, we avoid the need for multi-stage training or extra intermediate representations. Moreover, we eliminate the need for adversarial losses or lip expert supervision by directly adopting the DSM objective. 
For each training iteration, we sample a clean latent $z_0$ and noise $n \sim \mathcal{N}(0,\sigma^2)$, then set $z_t = z_0 + n$. 
Our loss is:
\begin{small}
\begin{align}
\setlength{\abovedisplayskip}{0pt}
\mathcal{L}
=
\mathbb{E}_{\substack{z_0,\,t\\n\,\sim\,\mathcal{N}(0,\sigma^2)}}
\Bigl[
  \lambda_{\sigma}\,
  \bigl\|\,
    \Phi_\theta\bigl(z_t,\,t,\,\mathcal{I}_r,\,a,\,V_m;\,\sigma\bigr)
    % \;-\;
    \!\! - \!\!
    z_0
  \bigr\|_2^2
\Bigr].
\label{eq:ours_dsm}
\setlength{\belowdisplayskip}{0pt}
\end{align}
\end{small}

\cref{fig:overview} illustrates the overall pipeline of our method.
Building on insights from our experiments regarding the relative strength of each condition, 
we first present in \S3.2 our editing control module, 
then introduce the identity preservation module in \S3.3, 
followed by the audio-driven module in \S3.4.
Finally, we discuss several key training details and hyper-parameter choices.
\begin{figure}[!htp]
\setlength{\abovecaptionskip}{0.cm}
\setlength{\belowcaptionskip}{0.cm}
\centering
\includegraphics[width=0.95\linewidth]{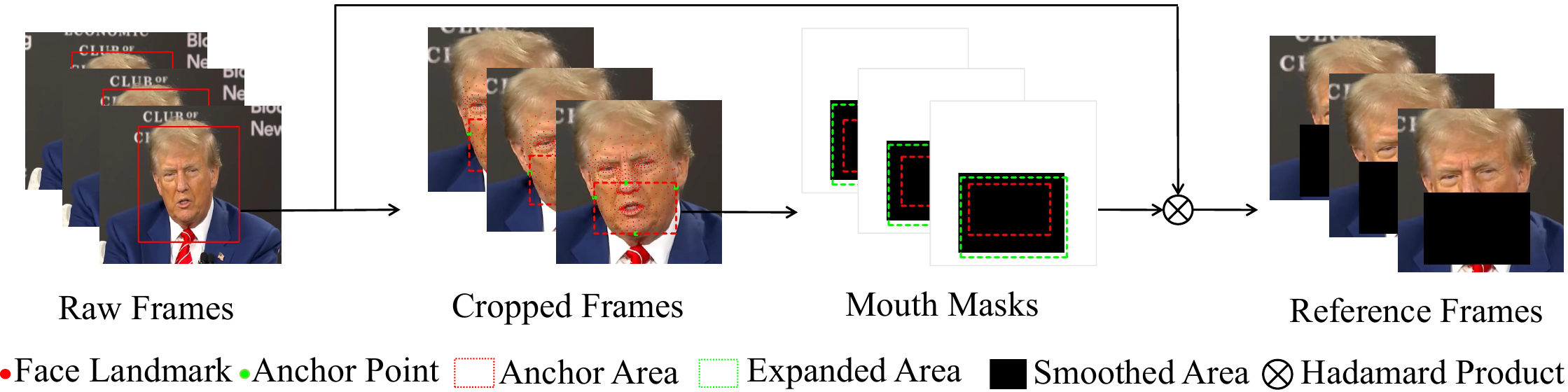}
\caption{Our adaptive masking strategy first determines the initial mask through detected landmarks, then obtains the final mask through expansion and smoothing, effectively preventing motion leakage.}
\label{fig:mask}
\end{figure}
% \vspace{-1cm}

\subsection{Editing Control}

The masked video sequence provides the strongest guidance signal in our framework, as the model only needs to synthesize the masked regions while preserving the rest. Previous studies \citep{mukhopadhyay2024diff2lip, yaman2023plug} have shown that lip motions often fail to align with input audio without SyncNet supervision, and we also observed this issue in our early experiments. Even with the lip region masked, initial experiments reveal that generated results maintain high similarity with the reference video regardless of the input audio, indicating the audio condition fails to guide lip motion synthesis effectively.

By comparing different masking strategies, we find that this motion leakage effect originates from the temporal context in the masked video sequence. With a low masking ratio, the model learns to infer lip movements from visible facial muscle patterns instead of audio features, as facial movements are highly correlated - visible facial regions can effectively indicate the movements of the masked lip region. Although this is not our desired optimization direction, it helps model convergence in the early training stages.

While increasing the mask coverage might be a solution, this approach alone is not optimal as it may remove essential facial features. To address this trade-off, we design an adaptive masking strategy that tracks head position across different head and lip poses. We opt for rectangular masks to avoid potential guidance from mask shapes, ensuring lip-surrounding pixels are masked and eliminating the influence of lower facial muscle motion patterns on lip movements. As shown in \cref{fig:mask}, the mask is generated using facial landmarks, with additional padding to ensure coverage of potential motion areas. To prevent motion leakage and maintain temporal stability, we apply a temporal smoothing process to the mask coordinates:
% \vspace{-0.8cm} 
\begin{equation}
\begin{split}
x_t^{\prime} &= \alpha x_t + (1-\alpha)x_{t+1} \\
y_t^{\prime} &= \alpha y_t + (1-\alpha)y_{t+1}
\end{split}
\end{equation}
where $\alpha=0.75$ provides a balance between temporal consistency and local accuracy. This smoothing process enhances mask movement stability and further prevents the model from inferring lip movements based on landmark motion patterns.

For implementation, we encode the masked video through the VAE encoder to obtain its latent representation. We then concatenate these latents with the noise latents along the channel dimension, forming an 8-channel input tensor. This design replaces the original first-frame concatenation in SVD while maintaining the same channel dimensionality, thereby providing rich temporal guidance for generation.
\subsection{Identity Preservation}

Existing methods \citep{chang2023magicdance,jiang2024loopy} typically employ ReferenceNet \citep{hu2024animate} for reference image encoding because it shares the same architecture with the denoising UNet and can be initialized with identical weights, enabling rapid condition-backbone alignment. However, directly adopting a stable video diffusion architecture would introduce substantial parameter redundancy.

Through experiments, we find that introducing complex architectures or large numbers of parameters not only increases computational overhead but also leads to excessive visual conditions. During our experiments, we discovered a spatial dependency issue: a strong identity condition influences the lip state in the reference image to dominate generation results.

To address computational efficiency and spatial dependency issues, we propose ID-Guider, an efficient encoding module comprising convolution layers. Before feeding the reference image into ID-Guider, we first concatenate the lip mask as an indicator channel with the reference image, then process it through a simple pure convolutional downsampler with channel dimensions $[32, 64, 128, 64]$ to align the input with the denoising UNet. Furthermore, compared to ReferenceNet, we remove computationally intensive modules such as 3D convolutions, retaining only 2D ResBlock modules with channel dimensions and layer counts consistent with stable video diffusion to ensure proper residual integration into the denoising UNet. We also eliminate upsample layers. Moreover, since we remove the timestep-related modules, ID-Guider does not need to recalculate following the denoising steps during inference. As a result, ID-Guider maintains only 98M parameters, reducing the parameter count by over 90\%. This significantly improves computational efficiency and reduces the influence of visual information from the reference image while maintaining strong identity preservation capabilities.

\subsection{Audio Guidance}
Audio signals, while relatively weak compared to visual conditions \citep{chen2024echomimic,tian2025emo}, provide essential features for driving lip movements. We employ Whisper \citep{radford2023robust} as our audio feature extractor for its robust audio representation capabilities. Following common practice in audio-driven synthesis, we align each video frame with a window of surrounding audio features $(x_{t-k}, ..., x_t, ..., x_{t+k})$ to capture temporal context, where $x_t$ denotes the audio feature at timestep $t$, and $k$ determines the temporal context range. For boundary frames, we apply zero-padding without additional processing.

To strengthen the influence of audio signals, we incorporate audio cross-attention into both the downsample and upsample modules of our denoising UNet. Specifically, the noisy latents act as queries, while the audio features serve as keys and values. Since these audio features already encode a contextual window, we focus on computing spatial attention that guides the lip region’s spatial distribution during generation.

\begin{table*}[!ht]
\caption{Quantitative comparison on HDTF~\citep{zhang2021flow} and AVASpeech~\citep{chaudhuri2018ava}. 
Arrows indicate whether higher ($\uparrow$) or lower ($\downarrow$) values are better. 
The best results are in \textbf{bold}. Notably, all baseline methods incorporate SyncNet~\citep{chung2017out} as lip expert supervision and LPIPS~\citep{zhang2018unreasonable} as an additional training loss, where the ground-truth (GT) score is computed using SyncNet on real videos. SayAnything outperforms these methods across most metrics without requiring such expert supervision.}
\label{tab:comparison}
\vskip 0.15in
\centering
\resizebox{\textwidth}{!}{%
\begin{tabular}{l c c c c c c c c c c c c}
\toprule
 & \multicolumn{6}{c}{\textbf{HDTF}} & \multicolumn{6}{c}{\textbf{AVASpeech}} \\
\cmidrule(lr){2-7}\cmidrule(lr){8-13}
\textbf{Method} & 
\textbf{FID}$\downarrow$ & 
\textbf{FVD}$\downarrow$ & 
\textbf{SSIM}$\uparrow$ & 
\textbf{PSNR}$\uparrow$ & 
\textbf{LPIPS}$\downarrow$ & 
\textbf{Sync-c}$\uparrow$ &
\textbf{FID}$\downarrow$ & 
\textbf{FVD}$\downarrow$ & 
\textbf{SSIM}$\uparrow$ & 
\textbf{PSNR}$\uparrow$ & 
\textbf{LPIPS}$\downarrow$ & 
\textbf{Sync-c}$\uparrow$ \\
\midrule
Diff2lip~\citep{mukhopadhyay2024diff2lip}   
& 39.11 & 647.42 & 0.5664 & 14.36 & 0.2336 & 3.55
& 54.80 & 578.05 & 0.5153 & 12.14 & 0.2649 & 2.09 \\
VideoRetalking~\citep{cheng2022videoretalking} 
& 15.48 & 328.33 & 0.0865 & 24.70 & 0.0528 & 7.60
& 18.93 & 354.27 & 0.8632 & 25.35 & 0.0550 & 7.98 \\
MuseTalk~\citep{zhang2024musetalk}       
& 8.42  & 209.79 & 0.8871 & 27.99 & 0.0297 & 6.39
& 15.76 & 331.71 & 0.8800 & 26.32 & 0.0414 & 6.46 \\
LatentSync~\citep{li2024latentsync}     
& 8.40  & 200.83 & 0.8870 & 27.67 & 0.0391 & \textbf{9.29}
& 13.91 & 234.95 & 0.8836 & 27.09 & 0.0451 & \textbf{8.31} \\
SayAnything           
& \textbf{5.68} & \textbf{197.71} & \textbf{0.9347} & \textbf{29.04} & \textbf{0.0239} & 7.23 
& \textbf{11.31} & \textbf{222.61} & \textbf{0.8881} & \textbf{27.12} & \textbf{0.0332} & 7.06 \\
\midrule
GT             
& --    & --     & --     & --    & --     & 8.58 
& --    & --     & --     & --    & --     & 7.84 \\
\bottomrule
\end{tabular}
}%
\vskip -0.1in
\end{table*}

\subsection{Training Strategy}

In our framework, multiple conditions are involved, including audio $a$, reference image $\mathcal{I}_r$, and masked video sequence $V_m$. Due to the inherently weaker correlation with lip movements, audio signals can be easily overwhelmed by visual features during training, leading to insufficient audio-driven control. Based on this observation, we adopt distinct masking strategies for the conditions during the training process. First, $a$ has a 5\% probability of being masked to zero, while $\mathcal{I}_r$ has a 15\% probability, and the masking of $a$ always triggers the masking of $\mathcal{I}_r$ to ensure audio-only generation scenarios. Second, all masking operations are applied to the fused features in the latent space by setting them to zero. The masked video sequence $V_m$ remains a fixed input condition without masking.

\section{Experiments}
\subsection{Experimental Settings}

\paragraph{Datasets.}
We utilize four public datasets for training: AVASpeech~\citep{chaudhuri2018ava}, HDTF~\citep{zhang2021flow}, VFHQ~\citep{wang2022vfhq}, and MultiTalk~\citep{sung2024multitalk}. AVASpeech (45 hours) was originally designed for speech activity detection and includes a portion of noisy audio segments. HDTF comprises 362 high-definition (HD) videos at resolutions mostly between 720p and 1080p. VFHQ contains over 16,000 high-fidelity video clips from diverse interview scenarios. Finally, MultiTalk is a multilingual dataset for 3D talking head generation; it spans 20 languages and 423 hours of video content. After filtering out videos with low face resolution, incomplete head regions, and audio-visual misalignment, we obtain approximately 600 hours of curated training data. For evaluation, we randomly sample 30 videos from the test sets of HDTF and AVASpeech datasets.

\paragraph{Implementation Details.}
We train our model on 8 NVIDIA H800 GPUs with a batch size of 16. The model is initialized with SVD weights~\citep{blattmann2023stable} and trained for 200k steps using AdamW optimizer~\citep{loshchilov2017decoupled} with a fixed learning rate of 6e-5. For training, video clips are processed at 25 fps with a resolution of 320×320 pixels, and each sequence consists of 16 frames. Audio inputs are resampled to 16 kHz. The reference frame for each sequence is randomly sampled from the corresponding complete video.

\paragraph{Evaluation Metrics.}
Our evaluation framework consists of three key aspects: (1) Visual fidelity: We employ FID~\citep{heusel2017gans} to evaluate the quality of generated frames, particularly focusing on identity preservation and visual details. SSIM and PSNR provide complementary measurements of reconstruction accuracy, while LPIPS~\citep{zhang2018unreasonable} captures perceptual similarities. (2) Temporal coherence: We adopt FVD~\citep{unterthiner2018towards} to assess video-level quality and motion consistency. (3) Audio-visual synchronization: The SyncNet confidence score~\citep{chung2017out} quantifies the accuracy of lip movements relative to audio input.

\subsection{Comparisons}
\begin{figure*}[!htp]
    \centering
 \subfloat[]{
 \includegraphics[width=0.525\textwidth]{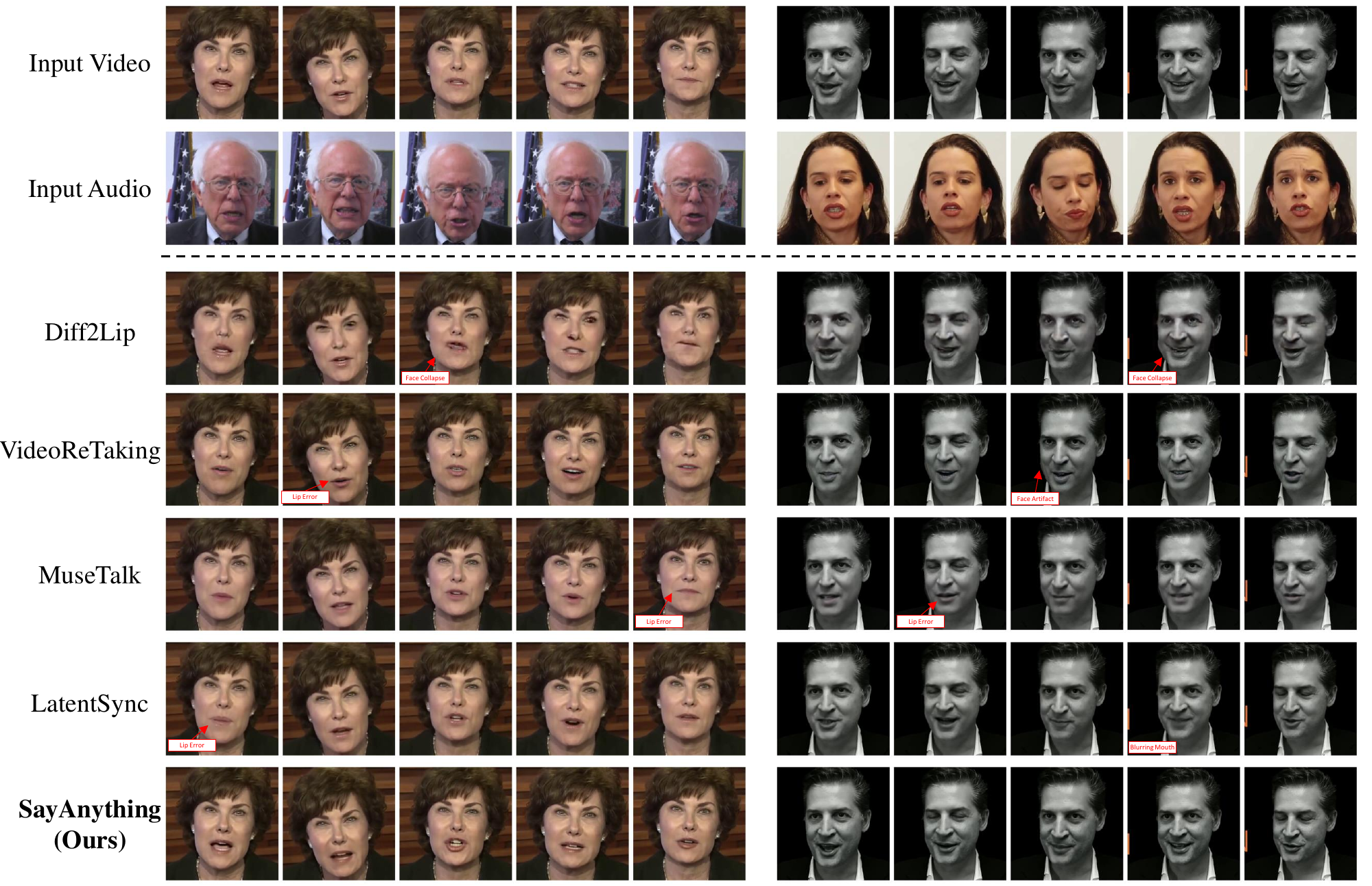}
 \label{fig:compare1}
 }
  \subfloat[]{
 \includegraphics[width=0.46\textwidth]{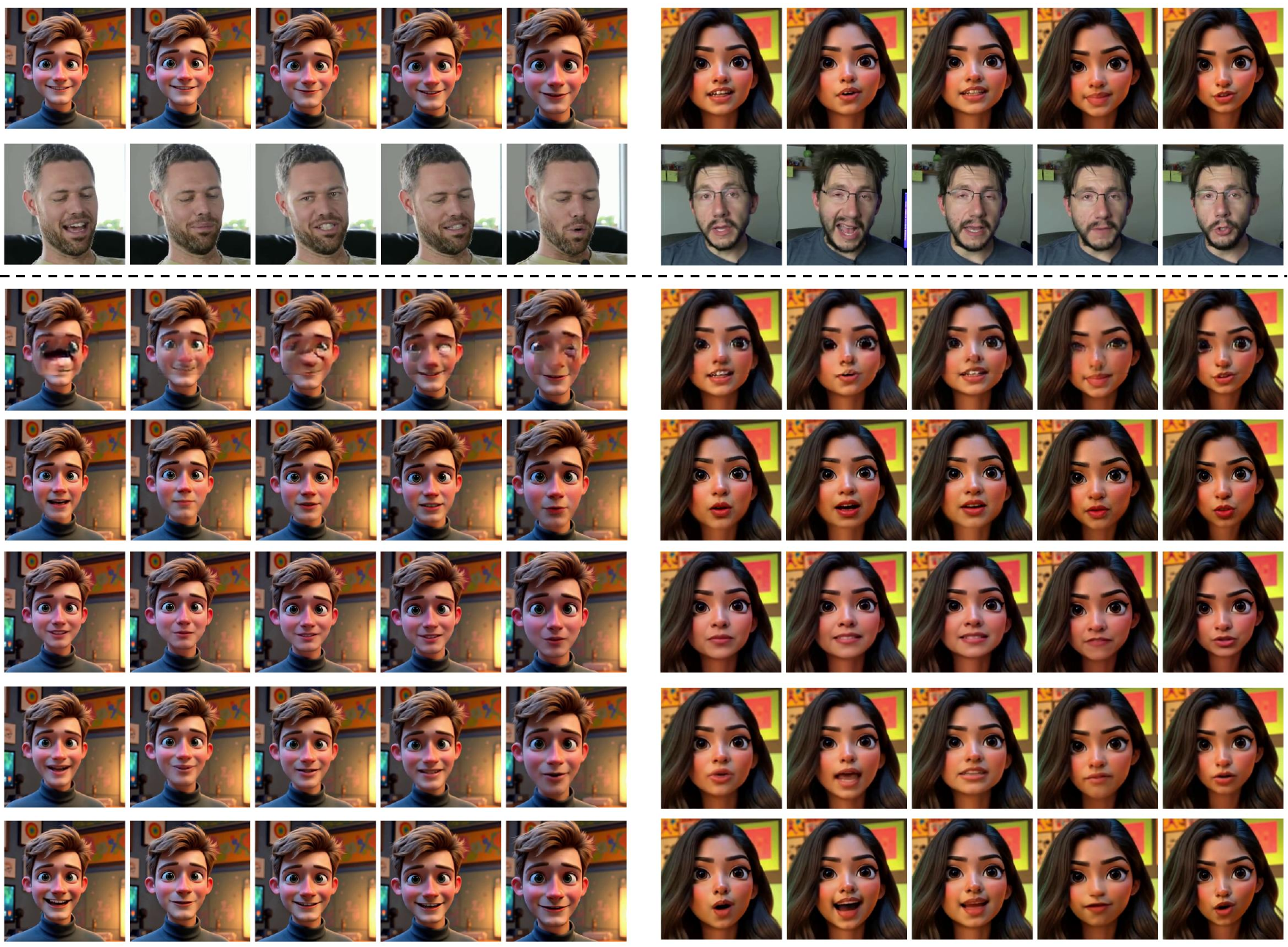}
 \label{fig:compare2_v2}
 }
    \caption{Qualitative comparisons with SOTA diffusion-based lip-sync methods~\citep{mukhopadhyay2024diff2lip,zhang2024musetalk,li2024latentsync,cheng2022videoretalking}. The first row demonstrates the original input video, and the second row is the video from which we extracted the audio as input, the video can be regarded as the target lip movements. Rows 3 - 7 display the lip-synced videos.  (a) Two cases in the cross-sex and ID generation setting.  (b) Two cases in the animate settings. Our method can generate more federal visual features like driven animators while others tend to generate fake features which are more realistic.
}
    \label{fig:compare}
\end{figure*}

\paragraph{Qualitative and Quantitative Comparison.}
We compare our method with state-of-the-art diffusion-based approaches that provide both inference code and pre-trained weights. Specifically, we consider: (1) LatentSync~\citep{li2024latentsync}, which utilizes an optimized SyncNet for additional supervision; (2) Diff2lip~\citep{mukhopadhyay2024diff2lip}, which directly generates the lower half of frames with adversarial and sync losses; (3) MuseTalk~\citep{zhang2024musetalk}, which builds upon the stable diffusion framework; and (4) VideoRetalking~\citep{cheng2022videoretalking}, which employs a three-stage network architecture.

As shown in \cref{tab:comparison}, our method demonstrates consistent improvements across most metrics on both datasets. While our method shows lower Sync-c scores~\citep{chung2017out}, this metric is computed by the original SyncNet model. Notably, all four baseline methods incorporate SyncNet as an additional supervision signal during training, with LatentSync employing an optimized version that produces scores exceeding ground truth values. Our results in visual quality, identity preservation, temporal consistency and competitive lip synchronization demonstrate the superiority of SayAnything.

\begin{figure}[!htp]
    \centering
    \includegraphics[width=0.45\textwidth]{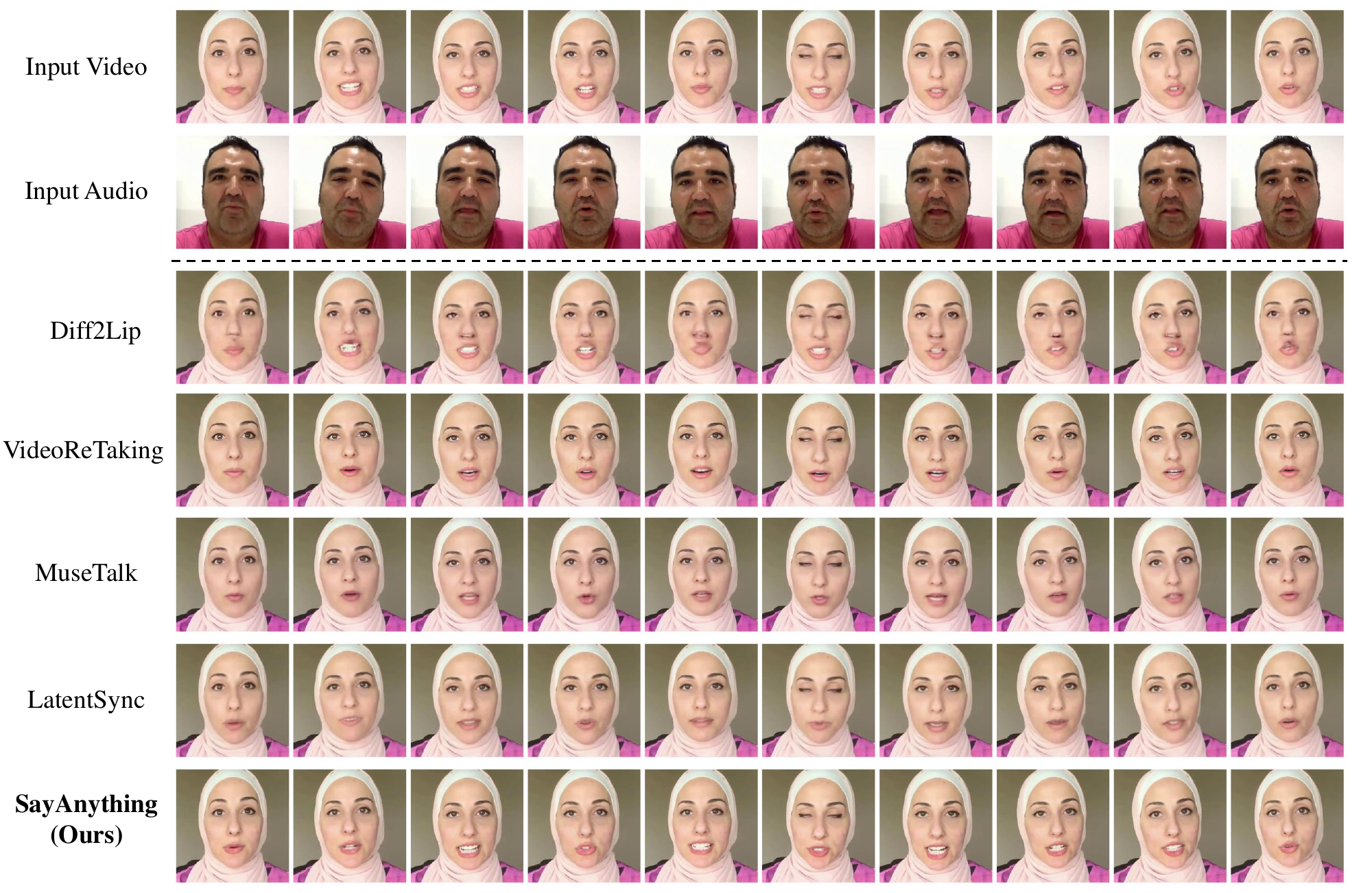}
    \caption{Qualitative comparison of lip motion dynamics and tooth rendering. Our method demonstrates clearer and more consistent teeth as well as more flexible lip movements.}
    \label{fig:compare-teeth}
    % \vspace{-1cm}
\end{figure}

As illustrated in \Cref{fig:compare}, existing methods exhibit various limitations: Diff2Lip~\citep{mukhopadhyay2024diff2lip} generates corrupted facial appearance, LatentSync~\citep{li2024latentsync} produces incoherent lip movements, MuseTalk~\citep{zhang2024musetalk} struggles with identity preservation, and VideoRetalking~\citep{cheng2022videoretalking} shows limited temporal consistency. Influenced by SyncNet's visual priors, these methods not only transform animated characters' lip shapes into life-like ones, compromising identity preservation, but also tend to generate conservative lip movements. In contrast, SayAnything maintains consistent identity preservation while generating more dynamic yet natural lip movements with high-quality teeth rendering and stable temporal coherence. Previous studies~\citep{jiang2024loopy,yaman2023plug} and comparisons of different methods~\citep{mukhopadhyay2024diff2lip,zhang2024musetalk,li2024latentsync,cheng2022videoretalking} in \cref{fig:compare-teeth} further demonstrate this phenomenon: SyncNet evaluation is unstable and favors conservative lip movements. Although our method shows significant improvements in visual metrics and lip motion dynamics, such more dynamic lip movements adversely affect the quantitative evaluation of lip synchronization.

\begin{table}[!htp]
\vspace{-0.5cm}
\caption{User preference rates (\%) across different scenarios. 
Users select one preferred method per test case. 
User study results across four video template categories 
Users upload their audio for testing. 
Results from 500 users show that SayAnything achieves the highest preference rates across all scenarios.}
\label{tab:userstudy}
\vskip 0.15in
\centering
\resizebox{\columnwidth}{!}{%
\begin{tabular}{lcccc}
\toprule
Method & Virtual & Cartoon & Life & News \\
\midrule
Diff2Lip~\citep{mukhopadhyay2024diff2lip} & 0.2 & 0.4 & 5.2 & 3.4 \\
VideoRetalking~\citep{cheng2022videoretalking} & 1.4 & 1.2 & 7.4 & 6.0 \\
MuseTalk~\citep{zhang2024musetalk}  & 2.2 & 2.2 & 9.4 & 11.4 \\
LatentSync~\citep{li2024latentsync} & 19.4 & 21.4 & 30.2 & 29.4 \\
SayAnything & \textbf{76.8} & \textbf{74.8} & \textbf{48.8} & \textbf{49.8} \\
\bottomrule
\end{tabular}%
}
% \vspace{-0.5cm}
\vskip -0.1in
\end{table}

\begin{table*}[t]
\caption{Ablation studies to validate the effectiveness of SayAnything modules. Video Fusion refers to encoding masked video through ID-Guider while processing reference images following stable video diffusion, concatenated with noise latent. VAE Feature indicates pre-encoding reference image through VAE. w/o Mask represents using larger fixed masks for videos rather than adaptive masking. w/o Cfg denotes the removal of the condition masking strategy during training.}
\label{tab:ablation}
\centering
\resizebox{\textwidth}{!}{%
\begin{tabular}{l c c c c c c c c c c c c}
\toprule
 & \multicolumn{6}{c}{\textbf{HDTF}} & \multicolumn{6}{c}{\textbf{AVASpeech}} \\
\cmidrule(lr){2-7}\cmidrule(lr){8-13}
\textbf{Method} &
\textbf{FID}\(\downarrow\) & 
\textbf{FVD}\(\downarrow\) & 
\textbf{SSIM}\(\uparrow\) & 
\textbf{PSNR}\(\uparrow\) & 
\textbf{LPIPS}\(\downarrow\) & 
\textbf{Sync-c}\(\uparrow\) &
\textbf{FID}\(\downarrow\) & 
\textbf{FVD}\(\downarrow\) & 
\textbf{SSIM}\(\uparrow\) & 
\textbf{PSNR}\(\uparrow\) & 
\textbf{LPIPS}\(\downarrow\) & 
\textbf{Sync-c}\(\uparrow\) \\
\midrule
Video Fusion &
20.14 & 247.73 & 0.8589 & 23.52 & 0.0504 & 6.09 &
20.32 & 280.65 & 0.8680 & 24.63 & 0.0497 & 6.09 \\
VAE Feature &
20.13 & 244.42 & 0.8569 & 23.20 & 0.0516 & 5.88 &
21.24 & 324.96 & 0.8642 & 23.81 & 0.0518 & 5.56 \\
w/o Mask &
16.19 & 299.44 & 0.8676 & 24.50 & 0.0483 & 6.54 &
17.80 & 299.85 & 0.8749 & 24.93 & 0.0497 & 6.30 \\
w/o Cfg &
14.58 & 255.29 & 0.8628 & 23.72 & 0.0510 & 5.14 &
18.65 & 257.97 & 0.8635 & 24.06 & 0.0523 & 5.69 \\
SayAnything &
\textbf{5.68} & \textbf{197.71} & \textbf{0.9347} & \textbf{29.04} & \textbf{0.0239} & \textbf{7.23} &
\textbf{11.31} & \textbf{222.61} & \textbf{0.8881} & \textbf{27.12} & \textbf{0.0332} & \textbf{7.06} \\
\bottomrule

\end{tabular}%
}
% \vspace{-0.5cm}
\end{table*}
\paragraph{User Study.}

We further evaluate SayAnything through a comprehensive user study. We provide 40 video templates that can be categorized into four types with broad application scenarios: virtual characters generated by AI models, cartoon characters rendered in styles similar to Disney animation films, news anchors in studio settings, and lifestyle scenarios featuring dynamic backgrounds and more significant head movements. Users can test with their own audio inputs, comparing results from all methods simultaneously and selecting the most preferred one. As shown in \cref{tab:userstudy}, SayAnything achieves the highest preference rates across all scenarios, demonstrating its versatility and superiority in handling diverse visual styles.

\subsection{Ablation Studies}

This section presents a comprehensive ablation analysis of SayAnything’s multi-modal fusion scheme and its key components. In earlier experiments, we attempted to use the ID-Guider module to encode the masked video, then replace the original first-frame image in stable video diffusion with the reference image. This seemed a natural approach, but we observed that the reference image ended up dominating the output. For instance, if the reference image had a wide-open mouth, the resulting video could no longer perform a closed-mouth action and instead remained open-mouthed throughout, which severely impairs lip synchronization accuracy. We believe this arises because the reference image gets replicated to align with the noise latent dimension, which originally helps lock in the first frame in the original model, but in our approach, it suppresses any subsequent mouth movement due to the spatial influence of the reference image.

\begin{figure}[!ht]
    % \vspace{-0.4cm}
    \centering
    \includegraphics[width=0.45\textwidth]{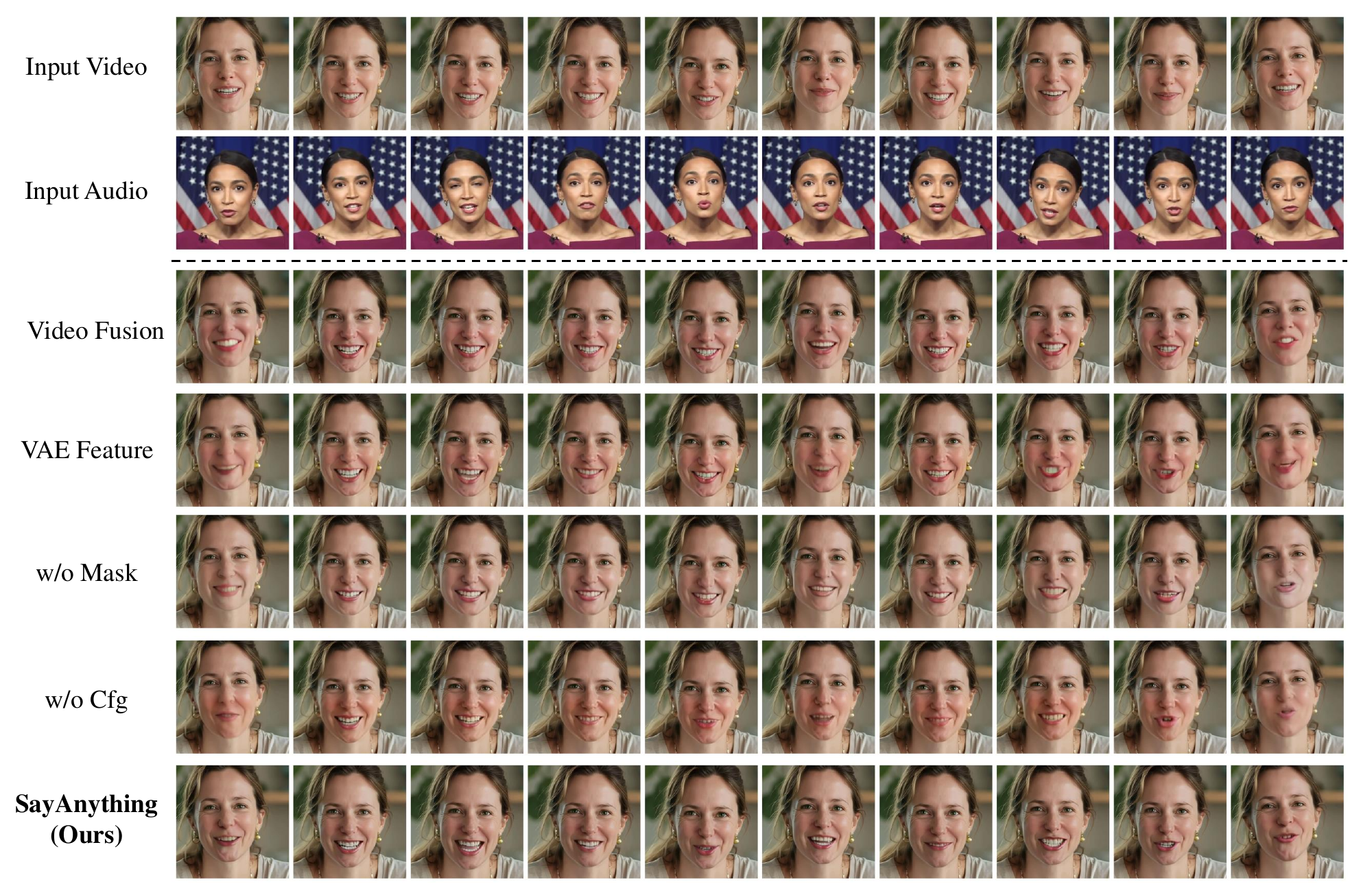}
    
    \caption{Ablation studies of our components in SayAnything. Video Fusion and VAE Feature significantly enhance reference image influence, limiting the range of lip movements. Larger fixed masks lead to colour shifts in masked regions and unnatural lip motions. Removing the condition masking strategy reduces visual quality. Zoom in for generated details.}
        
    \label{fig:ablastudy}
    % \vspace{-0.5cm}
\end{figure}

Likewise, using a VAE to compress the reference image into latent space and then input it to the ID-Guider produced a similar effect because it allowed the identity information to align with the denoising UNet from the earliest stage of training. Furthermore, applying a larger, fixed mask hampered consistent identity in the generated video and caused colour shifts between masked and unmasked regions, degrading visual quality. This finding suggests that our adaptive masking strategy effectively balances the editing region control with the audio-driven lip movements. Moreover, our conditional masking strategy further boosts the overall visual quality and stabilizes lip motions in the output video. \cref{fig:ablastudy} illustrates these phenomena and how our method improves them. We also evaluated these different training configurations on the HDTF~\citep{zhang2021flow} and AVASpeech~\citep{chaudhuri2018ava} datasets; as shown in \cref{tab:ablation}, our final approach’s key components demonstrate both strong effectiveness and well-grounded design in the fusion scheme.

\subsection{Generalization Capability}

\begin{figure}[!htp]
    \centering
    \setlength{\abovecaptionskip}{0.cm}
    \setlength{\belowcaptionskip}{0.cm}
    \includegraphics[width=0.45\textwidth]{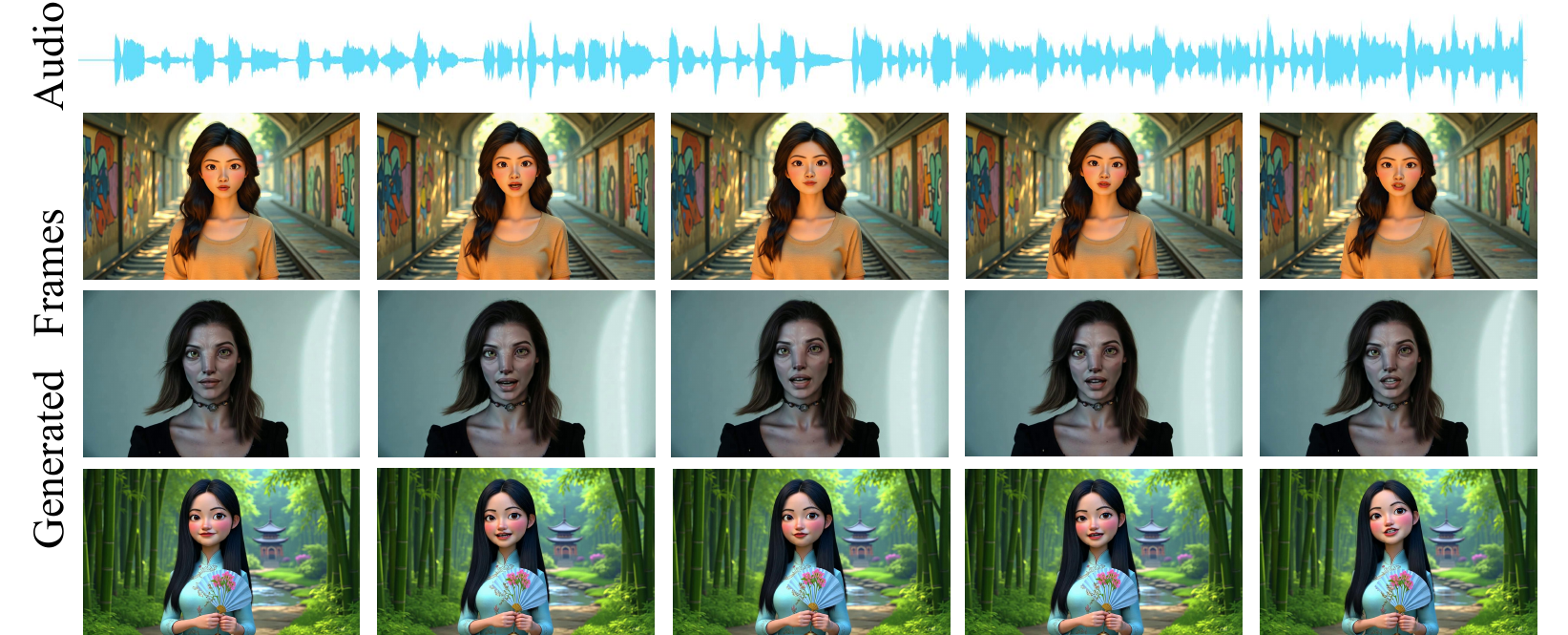}
    \caption{Visualization of pixar and virtual character videos generated by SayAnything.}
    \label{fig:gen}
    % \vspace{-0.3cm}
\end{figure}

Notably, SayAnything generalizes to out-of-domain video inputs without any additional fine-tuning. We validate this capability through our user study. As shown in \cref{fig:gen}, our method demonstrates zero-shot generalization to diverse animation styles while maintaining natural lip movements and tooth consistency. Additional visual results are provided in the supplementary materials.

\section{Conclusion}

We present SayAnything, an end-to-end video diffusion framework for audio-driven lip synchronization. Our method achieves natural lip movements with consistent teeth rendering and larger motion dynamics while maintaining identity preservation. Through a unified condition fusion scheme, SayAnything effectively balances audio-visual conditions and generates high-quality results without requiring additional supervision signals. The efficiency, innovation, and broad applicability of SayAnything make it promising for practical applications in lip synchronization.

% \begin{figure}[!htp]
%     \centering
%     \includegraphics[width=0.5\textwidth]{icml2024/figures/compare3.pdf}
%     \caption{Generated results within 5s. The first row demonstrates the original input video, and the second row is the video from which we extracted the audio as input, the video can be regarded as the target lip movements. Rows 3 - 7 display the lip-synced videos. It  demonstrates that our method can produce clearer and more consistent teeth as well as more flexible lip movements, while other methods are more influenced by the original video.}
%     \label{fig:compare-teeth}
% \end{figure}

\section*{Impact Statement}

The proposed SayAnything for audio-driven lip synchronization has significant potential applications across multiple domains. It can enhance digital content creation by enabling high-quality dubbing in different languages, facilitating international content distribution, and cross-cultural communication. The technology can improve virtual communication experiences through more natural and expressive virtual avatars in video conferencing and online education platforms. In the entertainment industry, our method can streamline the production of animated content and virtual characters by providing realistic lip movements synchronized with audio, benefiting film production, gaming, and digital media creation. Our experiments demonstrate that the model generalizes well across different scenarios, from actual human subjects to virtual characters and animated styles, enabling broad applications without domain-specific fine-tuning.

Potential Negative Social Impact: We acknowledge that this technology could potentially be misused for creating deceptive content, such as deepfake videos with manipulated speech. To address these concerns, we recommend: (1) implementing robust watermarking techniques to trace the origin of generated contents; (2) developing detection methods to identify AI-generated videos; (3) establishing clear guidelines and ethical frameworks for the deployment of such technologies. We emphasize the importance of collaborative efforts between researchers, industry practitioners, and policymakers to ensure the responsible development and application of audio-driven video synthesis technologies.

\bibliography{
    example_paper
}
\bibliographystyle{icml2025}

% %%%%%%%%%%%%%%%%%%%%%%%%%%%%%%%%%%%%%%%%%%%%%%%%%%%%%%%%%%%%%%%%%%%%%%%%%%%%%%%
% %%%%%%%%%%%%%%%%%%%%%%%%%%%%%%%%%%%%%%%%%%%%%%%%%%%%%%%%%%%%%%%%%%%%%%%%%%%%%%%
% % APPENDIX
% %%%%%%%%%%%%%%%%%%%%%%%%%%%%%%%%%%%%%%%%%%%%%%%%%%%%%%%%%%%%%%%%%%%%%%%%%%%%%%%
% %%%%%%%%%%%%%%%%%%%%%%%%%%%%%%%%%%%%%%%%%%%%%%%%%%%%%%%%%%%%%%%%%%%%%%%%%%%%%%%
% \newpage
\clearpage
\appendix

\section{Dataset Construction}

We collect and process four public datasets, AVASpeech \citep{chaudhuri2018ava}, HDTF ~\citep{zhang2021flow}, MultiTalk \citep{sung2024multitalk}, and VFHQ \citep{wang2022vfhq}. to build our training corpus. 
Among these, \emph{AVASpeech} is the largest and includes some noisy audio segments. 
\emph{HDTF} and \emph{VFHQ} contain predominantly high-definition (HD) video, thus providing detailed visual information suitable for our task. 
In the following, we outline our data filtering and pre-processing steps:
\vspace{-0.3cm}
\paragraph{Face Detection and Resolution Check.}
We employ YOLOv5 to detect faces in each video, retaining only the largest bounding box per frame. 
Any video whose face region never exceeds $228\times228$ pixels is discarded, removing clips with insufficient facial detail.
\vspace{-0.3cm}
\paragraph{Quality Filtering.}
Following VFHQ, we adopt HyperIQA~\citep{9156687} to remove videos exhibiting low clarity. 
Since most videos in our dataset share relatively consistent scenes, we additionally analyze frame-to-frame movements of the bounding box to exclude clips with excessive jitter or head motion. 
If the bounding box displacement across frames surpasses a threshold, we segment the video around those points to ensure each resulting clip is temporally coherent with a stable face region. 
Furthermore, we discard clips shorter than 2 seconds, which also effectively removes most multi-face scenes.
\vspace{-0.3cm}
\paragraph{Adaptive Mask Allowance.}
Due to our adaptive masking strategy (see Section~3.2 of the main text), we can tolerate up to eight consecutive frames without a detected face. 
In such cases, the mask is derived from contextual smoothing. 
Consequently, we adopt a lenient criterion for filtering side-face or partially occluded segments, aiming to maximize data utilization.
\vspace{-0.3cm}
\paragraph{Audio-Visual Alignment.}
Lastly, we remove instances where audio is clearly misaligned with the person on screen (e.g., voice from an off-screen speaker or background music instead of speech). 
Any clip with evidence of audio-video asynchrony or background-only audio is excluded to guarantee consistent lip and voice matching.

After applying all these steps, we obtain a curated dataset in which the speaker’s face is clearly visible and well-aligned with the corresponding audio. This final corpus spans a diverse range of speaking styles and resolutions, satisfying the requirements of our audio-driven lip synchronization framework.

\vspace{-0.3cm}
\section{Inference Procedure}
At test time, \emph{SayAnything} takes as input an audio clip and a video clip, which often differ in duration. We standardize to the \emph{audio} length: 
\begin{itemize}
    \item If the audio is shorter than the video, we simply truncate the video to match the audio length.
    \vspace{-0.3cm}
    \item If the audio is longer than the video, we concatenate the video in both forward and reverse orders, applying a smoothing function around the junction points to mitigate abrupt transitions.
\end{itemize}
\vspace{-0.3cm}
Once the video duration is matched to the audio, we detect and crop the face region in each frame. The bounding box is expanded by a certain proportion to avoid tight cropping, and its coordinates are smoothed across frames to prevent large jitter. We randomly select a reference image from the video frames to serve as the identity condition.

To generate the final lip-synced output, we adopt a \emph{segmented inference} approach, setting an overlap of four frames between adjacent segments. The guidance scale is set to 3.0, and we use 15 denoising steps, with the corresponding conditional CFG strategy as in training. After inference, we locate the face region in both the original and the generated videos. We dilate these bounding boxes slightly, then take their union as the final region into which the generated face is composited. This ensures a seamless integration of the lips and facial region in the output video.

\section{Limitations}

Despite several efficiency optimizations, our method's inference speed is still constrained by the diffusion process. As noted before, we trade off speed and quality by using 15 denoising steps, taking roughly 7 seconds on an RTX 4090 GPU to process 1 second of 25\,fps video. Moreover, \emph{SayAnything} has certain input limitations: for animated characters, it struggles with extended silent segments where the lips should remain static. Additionally, when faces are heavily occluded, the method often produces visually inconsistent results.

\section{Discussion on Motivation and Future Work}

We discuss potential research and engineering improvements as follows.
First, to further expand real-world applications, our method can be distilled into latent consistency models (LCMs)~\citep{luo2023latent}, substantially boosting inference efficiency.
Second, lip synchronization can exhibit a potential gap between audio signals and emotional expressiveness; by extending the region of interest from the lips to the entire head or even the full body, one can incorporate additional posture-based features that convey richer emotional cues.
Such directions are promising for enhancing both the performance and expressiveness of audio-driven generation.

\section{Addtional Visualizations}

\begin{figure*}[!ht]
    \centering
    \includegraphics[width=0.95\textwidth]{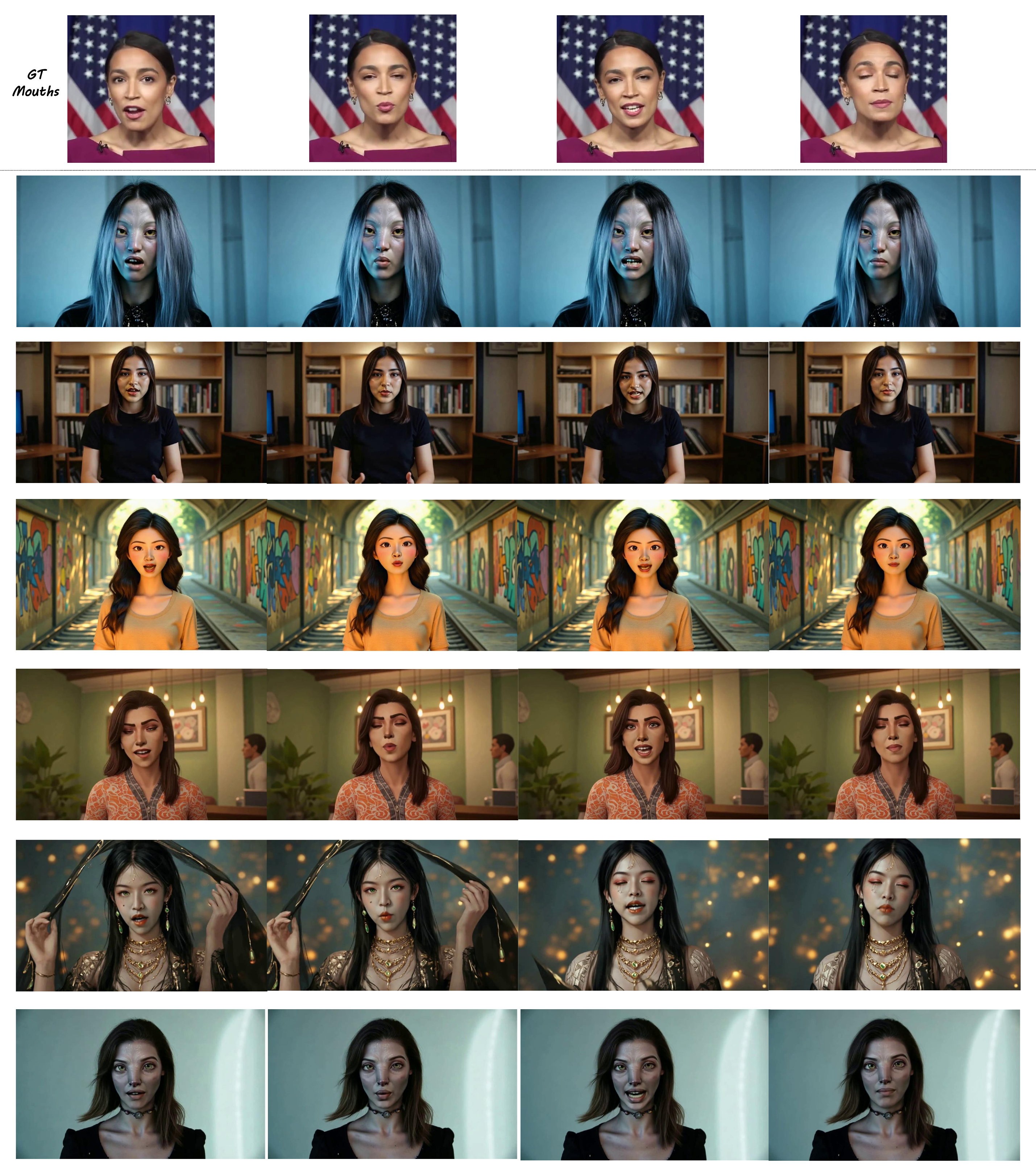}
    \caption{Lip synchronization results for different animated characters driven by the same audio segment. Our method demonstrates consistent lip motion patterns across various animation styles while preserving each character's unique visual characteristics.}
    \label{fig:add1}
\end{figure*}

In this section, we present additional generation results of SayAnything across various scenarios in \Cref{fig:add1}.

% \appendix
% \onecolumn

% You can have as much text here as you want. The main body must be at most $8$ pages long.
% For the final version, one more page can be added.
% If you want, you can use an appendix like this one.  

% The $\mathtt{\backslash onecolumn}$ command above can be kept in place if you prefer a one-column appendix, or can be removed if you prefer a two-column appendix.  Apart from this possible change, the style (font size, spacing, margins, page numbering, etc.) should be kept the same as the main body.
% %%%%%%%%%%%%%%%%%%%%%%%%%%%%%%%%%%%%%%%%%%%%%%%%%%%%%%%%%%%%%%%%%%%%%%%%%%%%%%%
% %%%%%%%%%%%%%%%%%%%%%%%%%%%%%%%%%%%%%%%%%%%%%%%%%%%%%%%%%%%%%%%%%%%%%%%%%%%%%%%
\end{sloppypar}
\end{document}